\documentclass[10pt,twocolumn,letterpaper]{article}
\usepackage{iccv}
\usepackage{times}
\usepackage{epsfig}
\usepackage{graphicx}
\usepackage{amsmath}
\usepackage{amssymb}
\usepackage{listings}

\usepackage{multirow}
\usepackage[dvipsnames]{xcolor}

\usepackage{rotating}

\usepackage[breaklinks=true,bookmarks=false]{hyperref}

\definecolor{mygreen}{rgb}{0,0.6,0}
\definecolor{mygray}{rgb}{0.5,0.5,0.5}
\definecolor{mymauve}{rgb}{0.58,0,0.82}

\iccvfinalcopy % *** Uncomment this line for the final submission

\def\iccvPaperID{8982} % *** Enter the ICCV Paper ID here

\ificcvfinal\fi

\begin{document}

%%%%%%%%% TITLE
\title{HybridAugment++: Unified Frequency Spectra Perturbations for Model Robustness}

\author{Mehmet Kerim Yucel$^{1}$ ~\quad Ramazan Gokberk Cinbis$^2$ ~\quad Pinar Duygulu$^1$ \quad 
\\
$^1$Hacettepe University, Graduate School of Science and Engineering~\quad  \\
$^2$Department of Computer Engineering, Middle East Technical University \\
{\tt\small mkerimyucel@gmail.com~\quad gcinbis@ceng.metu.edu.tr~\quad pinar@cs.hacettepe.edu.tr}
}

\maketitle
\ificcvfinal\thispagestyle{empty}\fi

%%%%%%%%% ABSTRACT
\begin{abstract}
Convolutional Neural Networks (CNN) are known to exhibit poor generalization performance under distribution shifts. Their generalization have been studied extensively, and one line of work approaches the problem from a frequency-centric perspective. These studies highlight the fact that humans and CNNs might focus on different frequency components of an image. First, inspired by these observations, we propose a simple yet effective data augmentation method \textit{HybridAugment} that reduces the reliance of CNNs on high-frequency components, and thus improves their robustness while keeping their clean accuracy high. Second, we propose \textit{HybridAugment++}, which is a hierarchical augmentation method that attempts to unify various frequency-spectrum augmentations. \textit{HybridAugment++} builds on \textit{HybridAugment}, and also reduces the reliance of CNNs on the amplitude component of images, and promotes phase information instead. This unification results in competitive to or better than state-of-the-art results on clean accuracy (CIFAR-10/100 and ImageNet), corruption benchmarks (ImageNet-C, CIFAR-10-C and CIFAR-100-C), adversarial robustness on CIFAR-10 and out-of-distribution detection on various datasets. \textit{HybridAugment} and \textit{HybridAugment++} are implemented in a few lines of code, does not require extra data, ensemble models or additional networks \footnote{Our code is available at \url{https://github.com/MKYucel/hybrid_augment}.}.

\end{abstract}

\section{Introduction} \label{intro_hybridaugment}

The last decade witnessed machine learning (ML) elevating many methods to new heights in various fields. Despite surpassing human performance in multiple tasks, the \textit{generalization} of these models are hampered by distribution shifts, such as adversarial examples \cite{szegedy2013intriguing}, common image corruptions \cite{hendrycks2019augmix} and out-of-distribution samples \cite{yang2021generalized}. Addressing these issues are of paramount importance to facilitate the wide-spread adoption of ML models in practical deployment, especially in safety-critical ones \cite{rosenberg2021adversarial,deng2020analysis}, where such distribution shifts are simply inevitable.

Distribution shift-induced performance drops signal a gap between how ML models and us humans perform perception. Several studies attempted to bridge, or at least understand, this gap from architecture \cite{Benz_2021_WACV,Yeo_2021_ICCV,Saikia_2021_ICCV} and training data \cite{hendrycks2019augmix,wang2021augmax,Lee_2020_CVPR_Workshops,chen2021amplitude,prime_aug,calian2022defending,Hendrycks_2022_CVPR} centric perspectives. An interesting perspective is built on the frequency spectra of the training data; convolutional neural networks (CNN) are shown to leverage high-frequency components that are invisible to humans \cite{wang2020high} and also shown to be reliant on the amplitude component, as opposed to the phase component humans favour \cite{chen2021amplitude}. Several studies leveraged frequency spectra insights to improve model robustness. These methods, however, either leverage cumbersome ensemble models \cite{Saikia_2021_ICCV}, formulate complex augmentation regimes \cite{frequencyaug_competitor_method,eccv2022_freqpaper} or focus on a single robustness venue \cite{long2022frequency,frequencyaug_competitor_method,eccv2022_freqpaper} rather than improving the broader robustness to various distribution shifts. Furthermore, it is imperative to preserve, if not improve, the clean accuracy levels of the model while improving its robustness.

\begin{figure*}[!t]
\begin{center}
      \includegraphics[width=1.\textwidth]{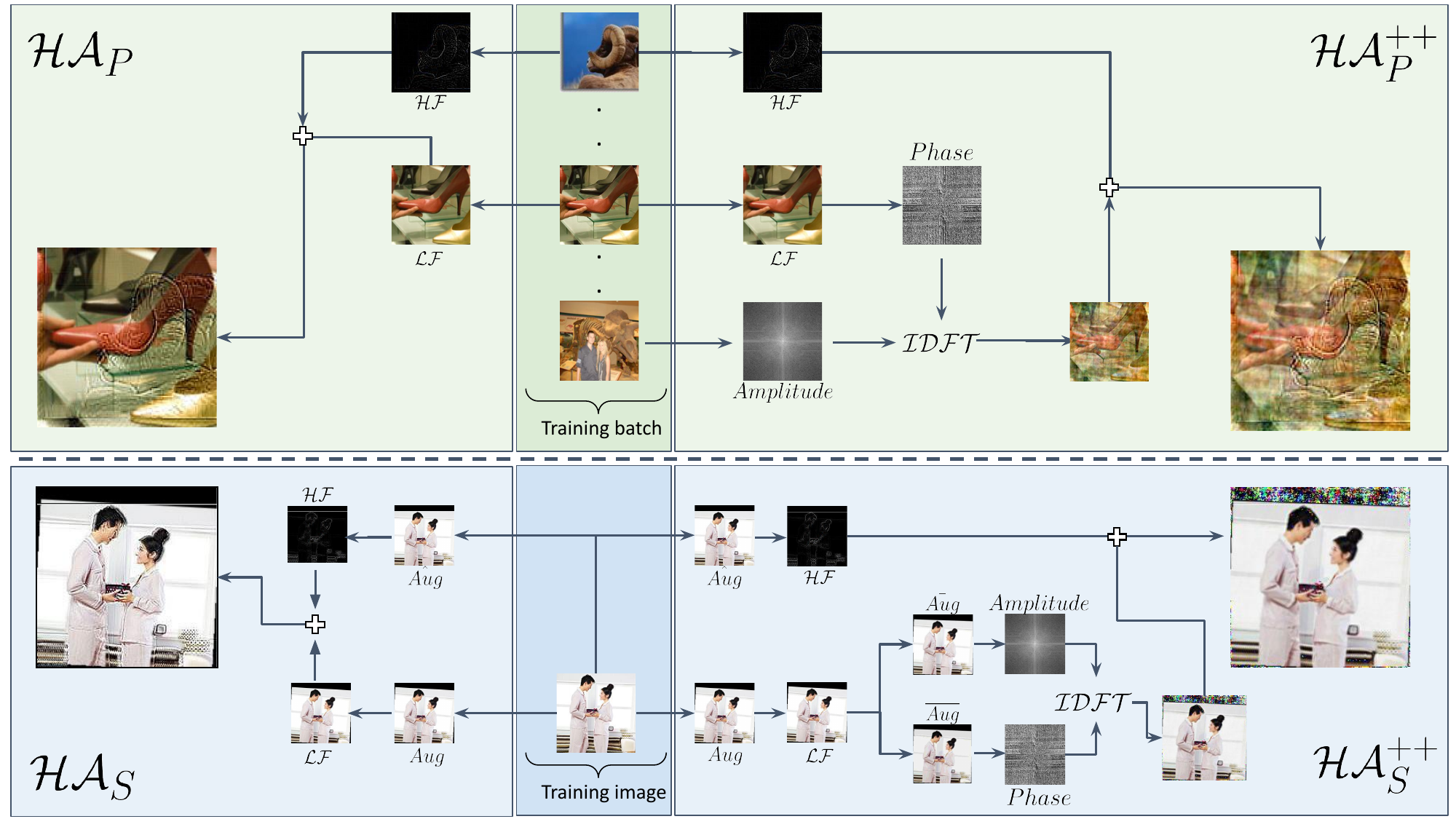}
    \caption{An overview of our methods \textit{HybridAugment} ($\mathcal{HA}$) and \textit{HybridAugment++} ($\mathcal{HA^{++}}$), and their single image ($\mathcal{_{S}}$) and paired ($\mathcal{_{P}}$) variants. $\mathcal{HA_{P}}$ combines the high-frequency ($\mathcal{HF}$) and low-frequency ($\mathcal{LF}$) contents of two randomly selected images, whereas $\mathcal{HA_{P}}^{++}$ combines the $\mathcal{HF}$ of one image with the amplitude and $\mathcal{LF}$-phase mixtures of two other images. Single image variants perform the same procedure, but based on different augmented versions of a single image.}        \label{fig:figure1_hybridaugment}
\end{center}
\vspace{-8mm}
\end{figure*}

Our work aims to improve the robustness of CNNs to various distribution shifts. Inspired by the frequency spectra based data augmentations,  we propose \textit{HybridAugment}, inspired from the well-known hybrid images \cite{oliva2006hybrid}. Based on the observation that the label information of images are predominantly related to the low-frequency components \cite{wang2020towards,li2022robust}, \textit{HybridAugment} simply swaps high-frequency and low-frequency components of randomly selected images in a batch, regardless of their class labels. This forces the network to focus on the low-frequency information of images and makes the models less reliant on the high-frequency information, which are often shown to be the root cause of robustness issues~\cite{wang2020towards}.  With virtually no training overhead, \textit{HybridAugment} 
improves the corruption robustness while preserving or improving the clean accuracy, and additionally induces adversarial robustness. 

Additionally, we set our eyes on jointly exploiting the contributions of frequency spectra augmentation methods while unifying them into a simpler, single augmentation regime. We then propose \textit{HybridAugment++}, which performs hierarchical perturbations in the frequency spectra. Exploiting the fact that the phase component carries most of the information in an image \cite{chen2021amplitude}, \textit{HybridAugment++} first decomposes images into high and low-frequency components, swaps the amplitude and phase of the low frequency component with another image, and then combines this augmented low-frequency information with the high-frequency component of a random image. Essentially, \textit{HybridAugment++} forces the models to rely on the phase and the low-frequency information. As a result, \textit{HybridAugment++} further improves adversarial and corruption robustness, while further improving the clean accuracy against several alternatives. See Figure \ref{fig:figure1_hybridaugment} for a diagram of our methods.

Our main contributions can be summarized as follows.

\begin{itemize}
\vspace{-2.5mm}
\item We propose \textit{HybridAugment}, a simple data augmentation method that helps models rely on low-frequency components of data samples. It is implemented in just three lines of code and has virtually no overhead.
\vspace{-2.5mm}
\item We extend \textit{HybridAugment} and propose \textit{HybridAugment++}, which performs hierarchical augmentations in frequency spectra to help models rely on low-frequency and phase components of images. 
\vspace{-2.5mm}
\item We show that \textit{HybridAugment} improves corruption robustness of multiple CNN models, while preserving (or improving) the clean accuracy. We additionally observe clear improvements in adversarial robustness over strong baselines via \textit{HybridAugment}.
\vspace{-2.5mm}
\item \textit{HybridAugment++} similarly outperforms many alternatives by further improving corruption and clean accuracies on multiple benchmark datasets, with additional gains in adversarial robustness.
\end{itemize}

\section{Related Work} \label{relwork_hybridaugment}

\noindent \textbf{Robust Generalization - Adversarial.} Adversarial ML has been studied intensively \cite{szegedy2013intriguing,yucel2020deep}, resulting into numerous attack \cite{szegedy2013intriguing,moosavi2016deepfool,goodfellow2014explaining} and defense \cite{madry2017towards,shaham2018defending,borkar2020defending,lu2017safetynet} methods borne out of an arms race that is still very much active. Notable attacks include FGSM \cite{goodfellow2014explaining}, DeepFool \cite{moosavi2016deepfool}, C\&W \cite{carlini2017towards} where AutoAttack \cite{croce2020reliable} is now a widely used attack for adversarial evaluation. The defense methods mainly diversify the training distribution with attacked images \cite{madry2017towards,zhang2020geometry}, purify the adversarial examples \cite{shaham2018defending,meng2017magnet} or detect whether an image is adversary or not \cite{xu2017feature,lu2017safetynet}.  

\noindent \textbf{Robust Generalization - Corruptions.} Common image corruptions might have various causes, and they occur more frequently than adversaries in practice. Numerous datasets simulating these effects have been released to facilitate standard evaluations \cite{hendrycks2019augmix,mu2019mnist,kar20223d,yucel2022robust}. The methods addressing corruption robustness can be largely divided into two; architecture-centric and data-centric methods. Architecture-centric methods include neural architecture search for robust architectures \cite{mok2021advrush}, focusing on subnets \cite{subnet_enhance}, rectifying batch normalization \cite{Benz_2021_WACV}, wavelet based layers \cite{li2020wavelet} and forming ensembles \cite{Saikia_2021_ICCV,Yeo_2021_ICCV}. The data-centric methods are arguably more prominent in the literature; adversarial training \cite{madry2017towards,adversarialtraining_corruption}, cascade augmentations \cite{hendrycks2019augmix,wang2021augmax}, augmentation networks \cite{ant_corruption,calian2022defending}, learned augmentation policies \cite{yin2019fourier}, shape-bias injection \cite{shapebias_corruption,Sun_2021_ICCV}, style augmentation \cite{geirhos2018imagenet}, fractals \cite{Hendrycks_2022_CVPR}, soft-edge driven image blending \cite{Lee_2020_CVPR_Workshops} and max-entropy image transformations \cite{prime_aug} are all shown to improve corruption robustness at varying degrees.

\noindent \textbf{Robust Generalization - Frequency Aspect.} The generalization of neural networks have been analysed extensively. Specificially, several frequency-centric studies show that CNNs tend to rely on high-frequency information ignored by human vision \cite{wang2020high}, or rely more on amplitude component than phase component humans tend to favour \cite{chen2021amplitude}. Models trained on high-pass filtered images are shown to have higher accuracy than the models trained on low-pass filtered images, although high-pass filtered images are just random noise to humans \cite{yin2019fourier}. Multiple studies confirm that models reliant on low-frequency components are more robust \cite{wang2020towards,li2022robust}. Interestingly, frequency analyses presents a different interpretation of the robustness-accuracy trade-off; many methods that improve clean accuracy force networks to rely on high-frequency components, which might sacrifice robustness \cite{wang2020high}. 

\noindent \textbf{Robust Generalization - Frequency-Centric Methods.} A trade-off in frequency-based data augmentations is that one should not sacrifice the other; training on high-frequency augmentations can improve robustness to high-frequency corruptions, but tend to sacrifice the low-frequency corruption robustness or the clean accuracy \cite{Saikia_2021_ICCV,yin2019fourier,chan2022does}. Frequency-centric methods include biasing Jacobians \cite{chan2022does}, swapping phase and amplitude of random images \cite{chen2021amplitude}, perturbing phase and amplitude spectra along with consistency regularization \cite{frequencyaug_competitor_method}, frequency-band expert ensembles \cite{Saikia_2021_ICCV}, frequency-component swapping of same-class samples \cite{mukai2022improving} and wavelet-denoising layers \cite{li2020wavelet}. Note that there is a considerable literature on frequency-centric adversarial attacks, but we primarily focus on methods improving robustness.

A similar work is \cite{mukai2022improving}, where hybrid-image based augmentation is proposed. We have, however, several key advantages; we i) lift the restriction of sampling from same classes for augmentation, ii) propose both single and paired variants, leading to a significantly more diverse training distribution, iii) present \textit{HybridAugment++} that performs phase/amplitude swap specifically in low-frequency components and iv) report improvements on corruption and adversarial robustness, as well as clean accuracy on multiple benchmark datasets (CIFAR-10/100, ImageNet). Note that other methods either train with ImageNet-C corruptions \cite{Saikia_2021_ICCV}, report only corruption results \cite{frequencyaug_competitor_method}, rely on external data \cite{Hendrycks_2022_CVPR} or models \cite{calian2022defending}. Our methods, on the other hand, require no external models or data, and they can be plugged into existing pipelines easily due to their simplicity.

\section{Method} \label{method_hybridaugment}
In this section, we formally define the problem, motivate our work and then present our proposed techniques.

\subsection{Preliminaries}
Let $\mathcal{F}(x;W)$ be an image classification CNN trained on the training set $\mathcal{T}_\text{train} = (x_{i}, y_{i})^{N}_{i=1}$  with $N$ samples, where $x$ and $y$ correspond to images and labels. The clean accuracy (CA) of $\mathcal{F}(x;W)$ is formally defined as its accuracy over a clean test set $\mathcal{T}_\text{test} = (x_{j}, y_{j})^{M}_{j=1}$. Assume two operators ${A}(\cdot)$ and ${C}(c, s)$ that adversarially attacks or corrupts a given set of images with the corruption category $c$ and severity $s$, respectively.  Let $A\mathcal{T}_\text{test}$ and $C\mathcal{T}_\text{test}$ be the adversarially attacked and corrupted versions of $\mathcal{T}_\text{test}$, and let $\mathcal{F}(x;W)$ have a robust accuracy (RA) on $A\mathcal{T}_\text{test}$ and a corruption accuracy (CRA) on $C\mathcal{T}_\text{test}$. 
The aim is to fit $\mathcal{F}(x;W)$ such that the model gains robustness (\ie. increased RA and CRA compared its the baseline version), while retaining (or improving) the clean accuracy of its baseline version trained without robustness concerns.

\noindent \textbf{What we know.} Our work builds on the following crucial observations: i) CNNs favour high-frequency content \cite{wang2020high}, ii) adversaries and corruptions often reside in high-frequency \cite{wang2020towards}, iii) images are dominated by low-frequency \cite{Saikia_2021_ICCV} and iv) models relying on low-frequency components are more robust \cite{li2022robust,wang2020towards}. The robustness-accuracy trade-off is visible; low-frequency reliant models are more robust, but tend to miss out on clean accuracy brought by the high-frequency components. 

\subsection{HybridAugment}
We hypothesize that a \textit{sweet spot} in the robustness-accuracy trade-off can be found. Unlike the \textit{hard} approaches that completely rule out the reliance on high-frequency components (i.e. low-pass filters), we propose to \textit{reduce} the reliance on them. To this end, we adopt a data augmentation approach that aims to diversify $\mathcal{T}_\text{train}$ by an operation $\mathcal{HA(\cdot)}$. Keeping the strong relation intact between labels and low-frequency content (i.e. labels come from low-frequency-component image), we propose to swap high and low-frequency components of images in a batch on-the-fly. Unlike \cite{mukai2022improving}, we \textit{do not} restrict the images to belong to the same class; this diversifies the training distribution even further while preserving the image semantics. We call this basic version of our approach \textit{HybridAugment}, which corresponds to: 
\begin{equation} \label{hybrid_augment_paired}
    \mathcal{HA_{P}}(x_{i}, x_{j}) = \mathcal{LF}(x_{i}) + \mathcal{HF}(x_{j})
\end{equation}
where $x_{i}$ is the input image and $x_{j}$ is a randomly sampled image from the whole training set, which we simply sample from the mini batch at each training iteration in practice. $\mathcal{HF}$ and $\mathcal{LF}$ operators select the high and low-frequency components of an input image, for which we use:
\begin{equation} \label{eq:cutoff}
\begin{split}
    \mathcal{LF}(x) = GaussBlur(x) \\
    \mathcal{HF}(x) = x - \mathcal{LF}(x)
    \end{split}
\end{equation}
where $GaussBlur$ is used as a low-pass filter. Note that a similar outcome is possible by using Discrete Fourier Transforms (DFT), swapping the frequency bands and then applying Inverse DFT (IDFT). We find the gaussian blur operation to be faster and better in practice.

Inspired from \cite{chen2021amplitude}, in addition to the image-pair scheme in Eq.~\ref{hybrid_augment_paired}, we propose a single image variant of \textit{HybridAugment}. In the single image variant, instead of combining two images, $x_i$ and $x_{j}$ are obtained by applying randomly sampled augmentations to a single image. The single image variant $\mathcal{HA_{S}}$ can therefore be defined as 
\begin{equation} \label{hybrid_augment_single}
    \mathcal{HA_{S}}(x_{i}) = \mathcal{LF}(Aug(x_{i})) + \mathcal{HF}(\hat{Aug}(x_{i}))
\end{equation}
where $Aug$ and $\hat{Aug}$ correspond to two sets of randomly sampled augmentation operations. Note that paired and single versions can work in tandem ($\mathcal{HA_{PS}}$), and actually outperform single or paired image versions.

\subsection{HybridAugment++}

The frequency analysis is a vast literature, however, two core aspects often stand out; frequency-band analysis (i.e. low, high) and the decomposition of signals into amplitude and phase. \textit{HybridAugment} covers the former and shows competitive results in various benchmarks (see Section \ref{sec:exp_hybridaugment}). The latter is investigated in $\mathcal{APR}$ \cite{chen2021amplitude}, where phase is shown to be the more relevant component for correct classification, and training models based on their phase labels and swapping amplitude components of images randomly lead to more robust models. Note that frequency-band and phase/amplitude discussions are arguably orthogonal, since frequency, phase and amplitude provide distinct characterizations of a signal: intuitively speaking, frequency, phase and amplitude can be seen as the separation of visual patterns in terms of scale, location and significance.

We hypothesize these two approaches can be complementary; a model reliant on low-frequency and spatial information (i.e. phase) can further improve robustness. Inspired by the successes of cascaded augmentation methods \cite{hendrycks2019augmix,wang2021augmax,calian2022defending}, we unify these two core aspects into a single, hierarchical augmentation method. We refer to this method as \textit{HybridAugment++} and define its paired version as:
\begin{equation}
  \mathcal{HA_{P}}^{++}(x_{i}, x_{j}, x_{z}) = \mathcal{APR_{P}}(\mathcal{LF}(x_{i}), x_{z}) + \mathcal{HF}(x_{j})
\end{equation}
where $x_{i}$, $x_{j}$ and $x_{z}$ are images sampled from the same batch. Here, $\mathcal{APR_{P}}$~\cite{chen2021amplitude} is defined as
\begin{equation}
    \mathcal{APR_{P}}(x_{i}, x_{z}) = \mathcal{IDFT}(A_{x_{z}} \otimes e^{i. P_{x_{i}}}) \\
\end{equation}
where $\otimes$ is element-wise multiplication, $A$ is the amplitude and $P$ is the phase component. Similar to $\mathcal{HA}$ and $\mathcal{APR}$, we also define a single-image version of \textit{HybridAugment++} as
\begin{equation}
 \mathcal{HA_{S}}^{++}(x_{i}) = \mathcal{APR_{S}}(\mathcal{LF}(Aug(x_{i}))) + \mathcal{HF}(\hat{Aug}(x_{i}))
\end{equation}
where $\mathcal{APR_{S}}$~\cite{chen2021amplitude} is defined as
\begin{equation}
\mathcal{APR_{S}}(x_{i}) = \mathcal{IDFT}\left(A_{\bar{Aug}(x_{i})} \otimes e^{i. P_{\overline{Aug}\left(x_{i}\right)}}\right)    
\end{equation}
where $Aug$, $\hat{Aug}$, $\bar{Aug}$ and $\overline{Aug}$ are different sets of randomly sampled augmentation operations. Note that we essentially propose a framework; one can use different single and paired image augmentations, either individually or together, and can still achieve competitive results (see ablations in Section \ref{sec:exp_hybridaugment}). There are also other alternatives, such as swapping phase/amplitude first and then performing $\mathcal{HA}$, but we observe poor performance in practice; dividing the phase component into frequency-bands is not interpretable as frequencies of the phase component are not well defined. The pseudo-code of our methods can be found in the supplementary material.

\section{Experimental Results} \label{sec:exp_hybridaugment}
In this section, we first describe our experimental setup, including the datasets, metrics, architectures and implementation details. We then present a discussion of the Gaussian kernel details, an important detail of the proposed schemes. We thoroughly evaluate the effectiveness of $\mathcal{HA}$ and $\mathcal{HA^{++}}$ in terms of three distribution shifts; common image corruptions, adversarial attacks, and out-of-distribution detection. We finalize with additional results and a discussion of the potential limitations.

\subsection{Experimental setup}

\noindent \textbf{Datasets.} We use CIFAR-10, CIFAR-100 \cite{krizhevsky2009learning} and ImageNet \cite{deng2009imagenet} for training. Both CIFAR datasets are formed of 50.000 training images with a size of 32$\times$32. ImageNet dataset contains around 1.2 million images of 1000 different classes. Corruption robustness is evaluated on the corrupted versions of the test splits of these datasets, which are CIFAR-10-C, CIFAR-100-C and ImageNet-C \cite{hendrycks2019benchmarking}. For each dataset, corruptions are simulated for 4 categories (noise, blur, weather, digital) with 15 corruption types, each with 5 severity levels.  For adversarial robustness, we use AutoAttack \cite{croce2020reliable} on CIFAR-10 test set. Out-of-distribution detection is evaluated on SVHN \cite{netzer2011reading}, LSUN \cite{yu2015lsun}, ImageNet and CIFAR-100, and their fixed versions \cite{tack2020csi}. 

\noindent \textbf{Evaluation metrics.} We report top-1 classification as clean accuracy. Adversarial robustness is evaluated with robust accuracy, which is the top-1 classification on adversarially attacked test sets. Corruption robustness is evaluated with Corruption Error (CE) $CE=\sum^{5}_{1}E_{c,s}^{F} / \sum^{5}_{1}E_{c,s}^{AlexNet}$. CE calculates the average error of the model $F$ on a corruption type, normalized by the corruption error of AlexNet \cite{krizhevsky2017imagenet}. CE is calculated for all 15 corruption types, and their average Mean Corruption Error (mCE) is used as the final robustness metric. Out-of-distribution detection is evaluated using the Area Under the Receiver Operating Characteristic Curve (AUROC) metric \cite{hendrycks2016baseline}.

\begin{figure}[!t]
\begin{center}
      \includegraphics[width=0.5\textwidth]{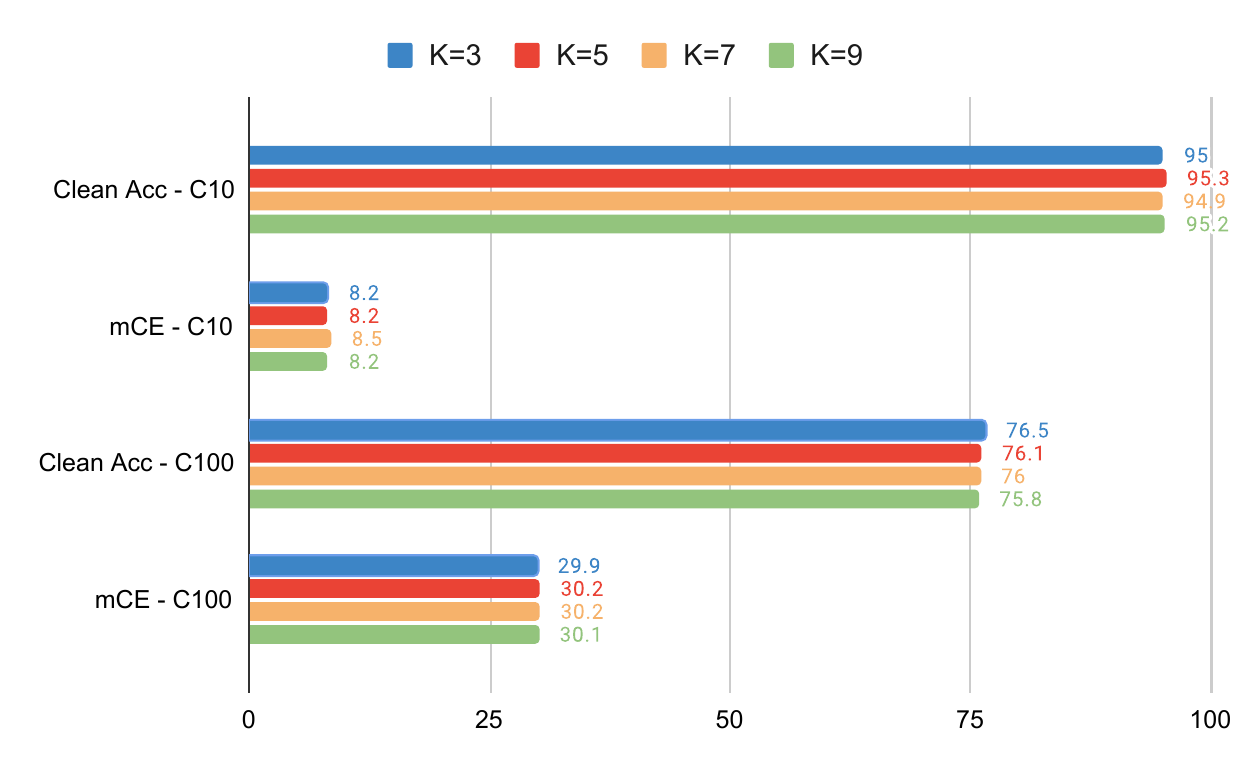}
      \includegraphics[width=0.5\textwidth]{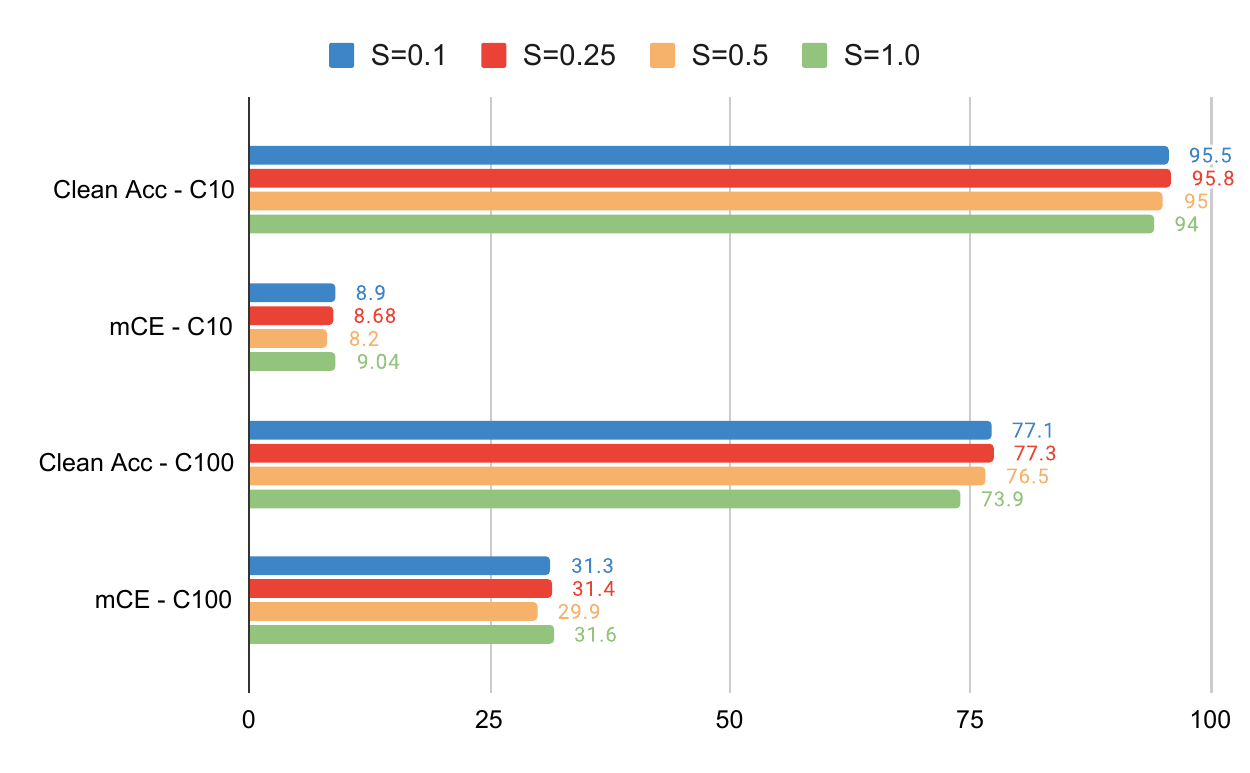}
      \vspace{-7mm}
    \caption{Clean accuracy $\uparrow$ and and mean corruption errors $\downarrow$ on CIFAR10/100, where different kernel sizes $\text{K}$ vs. a fixed standard deviation $\text{S}$ (top bar chart) and different standard deviations vs. a fixed kernel size are used for the blurring operation of Equation \ref{eq:cutoff}. }    
    \vspace{-8mm}
    \label{fig:blur_ablation}
\end{center}
\end{figure}

\noindent \textbf{Architectures.} We use architectures commonly used in the literature for a fair comparison; ResNeXT \cite{xie2017aggregated}, All-Convolutional Network \cite{springenberg2014striving}, DenseNet \cite{huang2017densely}, WideResNet \cite{zagoruyko2016wide} and ResNet18 \cite{he2016deep} are used in CIFAR-10 and CIFAR-100, whereas ResNet50 is used for ImageNet.

\noindent \textbf{Implementation details.} For CIFAR experiments, all architectures are trained for 200 epochs with SGD, where initial learning rate of 0.1 decays after every 60 epochs. We use the last checkpoints for evaluation and do not perform any hyperparameter tuning. Paired and single variants of $\mathcal{HA}$ and $\mathcal{HA^{++}}$ are applied in each iteration with probabilities 0.6 and 0.5, respectively. Standard training augmentations are random horizontal flips and cropping. When a single-image augmentation is used, the input image is augmented with $Aug$ randomly sampled among [\textit{rasterize, autocontrast, equalize, rotate, solarize, shear, translate}]. Note that these do not overlap with test corruptions. On ImageNet, we train for 100 epochs with SGD, where an initial learning rate of 0.1 is decayed every 30 epochs. Data augmentations and their probabilities are the same as above.

We use the same checkpoints for all evaluations; we do not train separate models for corruption and out-of-distribution detection. In adversarial analysis, for a fair comparison with \cite{chen2021amplitude}, we train our model with $\mathcal{HA}$ \& $\mathcal{HA^{++}}$ and FGSM adversarial training. We note that we use the labels of the low-frequency image as the ground-truth labels. We have tried using the high-frequency image labels instead, but this leads to severe degradation in overall performance, as expected. All models are trained with the cross-entropy loss, where the original and the augmented (with our method) batches are used to calculate the loss.

\subsection{Understanding the cut-off frequency} 

A key design choice is the cut-off frequency that defines $\mathcal{HF}$ and $\mathcal{LF}$ in Equation \ref{eq:cutoff}. Since we essentially define the cut-off frequency with a Gaussian blur operation, we have two hyperparameters; the size of the Gaussian kernel $\text{K}$ and its standard deviation $\text{S}$. Note that increasing both the kernel size and the standard deviation increases the blur strength, which eliminates increasingly higher frequencies (\ie higher cut-off frequency). We now evaluate the effects of these hyperparameters on both clean accuracy and mean corruption errors using $\mathcal{HA^{++}_{PS}}$, on both CIFAR-10 and CIFAR-100 using the ResNet18 architecture.

\noindent \textbf{Fixed standard deviation.} The effect of different $\text{K}$ values with fixed $\text{S}=0.5$ is shown in Figure \ref{fig:blur_ablation} top plot. $\text{K}=3$ provides the best trade-off here; it has the best clean accuracy and mCE on CIFAR100, whereas it shares the best mCE and has competitive clean accuracy on CIFAR10. 

\noindent \textbf{Fixed kernel size.} $\text{K}=3$ with different standard deviation values are shown in Figure \ref{fig:blur_ablation} bottom plot. The robustness-accuracy trade-off becomes more visible here; lower sigma values (\ie lower cut-off frequency) preserve more high-frequency content, and therefore have increasingly higher clean accuracy, but at the expense of degrading mCE. Note that further increasing the value $\text{S}$ is in contrast with this phenomena; if our method had only done frequency swapping (\ie $\mathcal{HA}$), then we could have expected a consistent trend, as shown in the literature \cite{li2022robust,wang2020towards}. However, $\mathcal{HA^{++}}$ also emphasizes the phase components, which results into a favourable behaviour where best results in mCE and clean accuracy can be obtained in the same cut-off frequency.

\noindent \textbf{The takeaway.} The results show that our hypothesis holds; we can find a sweet spot in the frequency spectrum where we can obtain favourable performance on both corrupted and clean images, given a careful selection of $\text{K}$ and $\text{S}$. A sound argument is that the optimality of these hyperparameters depends on the data; this is probably a correct assumption and can help tune the results further on other datasets. However, we use $K=3, S=0.5$ on all experiments across all architectures and datasets (including ImageNet), and show that we get solid improvements without any dataset-specific tuning.

\begin{table*}[]
\resizebox{\textwidth}{!}{%
\addtolength\tabcolsep{-3pt}
\begin{tabular}{l|c||ccc||cccc||ccc||ccc||ccc}
&  & \multicolumn{3}{c||}{Single-only} & \multicolumn{4}{c||}{Paired-only} & \multicolumn{3}{c||}{$\mathcal{APR_{P}}$\cite{chen2021amplitude} $with$} & \multicolumn{3}{c||}{$\mathcal{HA_{P}}$ $with$} & \multicolumn{3}{c}{$\mathcal{HA^{++}_{P}}$ $with$} \\ 
Method  & Orig &  $\mathcal{APR_{S}}$\cite{chen2021amplitude} & $\mathcal{HA_{S}}$ & $\mathcal{HA^{++}_{S}}$ & RFC\cite{mukai2022improving} & $\mathcal{APR_{P}}$ & $\mathcal{HA_{P}}$ & $\mathcal{HA^{++}_{P}}$ & $\mathcal{APR_{S}}$ & $\mathcal{HA_{S}}$ &  $\mathcal{HA^{++}_{S}}$&  $\mathcal{APR_{S}}$  & $\mathcal{HA_{S}}$ & $\mathcal{HA^{++}_{S}}$ & $\mathcal{APR_{S}}$ & $\mathcal{ HA_{S}}$ & $\mathcal{HA^{++}_{S}}$ \\ \hline
 AllConv & 30.8 & 14.8 & 16.8 & \underline{13.9} & 24.2 & 21.5 & 20.8 & \underline{16.7} & 11.5 &11.9 & \underline{11.2} & 11.5 & 12.0 & \underline{11.2} &10.9 & 10.9 & \textbf{10.7} \\
 DenseNet & 30.7  &12.3 & 15.0 & \underline{11.1} & 20.4 & 20.3 & 18.4 & \underline{14.2} & 10.3 & 10.6 & \underline{10.2} & 10.5 & 10.9 & \underline{10.2} & 10.1 & 10 & \textbf{9.5} \\
  WResNet & 26.9  & 10.6 & 13.6 & \underline{10.0} &18.3& 18.3 & 16.4 & \underline{13.2} & 9.1 & 9.2 & \underline{8.7} &9.4 & 9.9 & \underline{9.2} & 8.5 & 8.5 & \textbf{8.3} \\
  ResNeXt & 27.5  & 11.0 & 13.2 & \underline{9.99} & 19.2 &18.5 & 17.6 & \underline{13.2} & 9.1 & 9.3 & \underline{8.7}  & \underline{9.5} & 10.3 & \underline{9.5} &8.3 & 8.2 & \textbf{7.9} \\
  ResNet18 & 25.4  & 9.9 & 12.2 & \underline{9.34} &19.6& 17.0 & 18.3 & \underline{15.2} & 9.1 & 9.0 & \underline{8.5} &9.3 & 9.3 & \underline{9.0} &8.6 & 8.4 & \textbf{8.2} \\ \hline
\multicolumn{1}{l|}{Mean} & 28.3  & 11.7 & 14.1 & \underline{10.9} & 20.3 & 19.1 & 18.3 & \underline{14.5} & 9.8 & 9.9  & \underline{9.4} &10.0 & 10.4 & \underline{9.8} &9.2 & 9.2 & \textbf{8.9} \\ \hline
 AllConv & 56.4  & 39.8 & 43.0 & \underline{38.9} & 50.8 & 47.5 & 44.7 & \underline{41.7} & 35.9 & 35.9 & \underline{35.1}  &35.6 & 36.5 & \underline{34.8} & \textbf{34.4} & 34.6 & \textbf{34.4} \\
  DenseNet & 59.3  & 38.3 & 41.3 & \underline{37.3} & 52.5 &49.8 & 45.6 & \underline{41.8} & 35.8 & 36.3 & \underline{35.0}  & 34.8 & 36.1 & \underline{35.0} & 34.3 & 34.3 & \textbf{33.4} \\
  WResNet & 53.3 & 35.5 & 38.1 & \underline{33.9} & 47.4 & 44.7 & 43.1 & \underline{39.3} & 32.9 & 33.2 & \underline{31.9} & 32.9 & 34.2 & \underline{32.7} &31.5 & 31.4 & \textbf{31.2} \\
  ResNeXt & 53.4  & 33.7 & 35.6 & \underline{31.1} &46.4 & 44.2 & 41.2 & \underline{36.4} & 31.0 & 31.2 & \underline{29.9} & 30.6 & 31.5 & \underline{30.5} &30.5 & 29.0 & \textbf{28.8} \\
  ResNet18 & 51.2  & 33.0 & 35.6 & \underline{32.1} & 47.5 &49.2 & 45.5 & \underline{44.6} & 31.8 & 32.5 & \underline{31.2} & 32.2 & 31.8 & \underline{31.0} &30.3 & 30.4 & \textbf{29.9} \\ \hline
\multicolumn{1}{l|}{Mean} & 54.7  & 36.0 & 38.7 & \underline{34.6} & 48.9& 47.0 & 44.0 & \underline{40.7} & 33.4 & 33.8 & \underline{32.6}  & 33.3 & 34.0 & \underline{32.8} &32.2 & 31.9 & \textbf{31.5} \\ \hline
\end{tabular}%
}
\vspace{1mm}
\caption{Corruption robustness on CIFAR-10 (first 6 rows) and CIFAR-100 with various CNNs. Values show mCE, \textit{lower is better}. \underline{Underlined} scores are the best results within their respective group (i.e. single-only, paired-only, etc.). The overall best results are shown in \textbf{bold}. The table is divided into groups for easy comparison; single-only augmentation, paired-only augmentation and fixing one augmentation in paired variants while changing the single-image augmentation. }
\label{tab:cifar10_100_corruption}
\end{table*}

  % corruption robustness

% \usepackage{graphicx}
\begin{table*}[]
\resizebox{\textwidth}{!}{%
\addtolength\tabcolsep{-3pt}
\begin{tabular}{l|c||ccc||cccc||ccc||ccc||ccc}
&  & \multicolumn{3}{c||}{Single-only} & \multicolumn{4}{c||}{Paired-only} & \multicolumn{3}{c||}{$\mathcal{APR_{P}}$\cite{chen2021amplitude} $with$} & \multicolumn{3}{c||}{$\mathcal{HA_{P}}$ $with$} & \multicolumn{3}{c}{$\mathcal{HA^{++}_{P}}$ $with$} \\ 
Method  & Orig &  $\mathcal{APR_{S}}$\cite{chen2021amplitude} & $\mathcal{HA_{S}}$ & $\mathcal{HA^{++}_{S}}$ & RFC\cite{mukai2022improving} & $\mathcal{APR_{P}}$ & $\mathcal{HA_{P}}$ & $\mathcal{HA^{++}_{P}}$ & $\mathcal{APR_{S}}$ & $\mathcal{HA_{S}}$ &  $\mathcal{HA^{++}_{S}}$&  $\mathcal{APR_{S}}$  & $\mathcal{HA_{S}}$ & $\mathcal{HA^{++}_{S}}$ & $\mathcal{APR_{S}}$ & $\mathcal{ HA_{S}}$ & $\mathcal{HA^{++}_{S}}$ \\ \hline
AllConv & 93.9  & 93.5 & \underline{94.1} & 93.9 & 93.9 & \textbf{94.5} & 93.9 & 94.0 & \underline{94.3} & \underline{94.3} & \underline{94.3} & \textbf{94.5} & \textbf{94.5} & 94.4 & \textbf{94.5} & 94.4 & 94.3 \\
DenseNet & 94.2 & 94.9 & 94.7 & \underline{95.0} &93.6& \underline{95.0} & 93.1 & 93.2 & \textbf{95.2} & 95.1 & 95.1 &  94.7 & 95.0 & 94.9& 94.8 & \underline{95.0} & 94.8 \\
WResNet & 94.8 & 95.0 & 95.3 & \underline{95.4} & 93.0& \underline{95.2} & 93.2 & 92.0 & 95.7  & 95.4 & \textbf{95.8} & 95.4 & 95.5 & 95.2 & \underline{95.7} & 95.3 & 95.3 \\
ResNeXt & 95.7 & 95.5 & 95.3 & \underline{95.7} & 93.5 & \underline{95.5} & 93.5 & 92.9 & \textbf{96.1} & 95.6 & \textbf{96.1} & 95.4 &  95.2 & 95.1 & 95.6 & \underline{96.0} & 95.9 \\
ResNet18 & 92.2 & \textbf{95.6} & 95.5 & \textbf{95.6} &91.7& \underline{94.9} & 90.9 & 89.7 & 95.0 & 95.2 & \underline{95.4} & 95.4 & 95.4 & 95.1 & 95.0 & \underline{95.1} & 95.0 \\ \hline
Mean & 94.2  & 94.9 & 94.9 & \underline{95.1} & 93.0 & \underline{95.0} & 92.9 & 92.3 & 95.2 &95.1& \textbf{95.3} & \underline{95.1} & \underline{95.1} & 95.0 & 95.1 & \underline{95.2} &  95.1 \\ \hline
AllConv & 74.9 & 75.3 & 75.0 & \textbf{75.8} &\underline{75.3} & 74.8 & 74.1 & 74.7 & 75.2 & \underline{75.7} & 75.1 & 74.9 & \textbf{75.8} & 75.0 & 75.7 & \underline{75.6} & 75.2 \\
DenseNet & 71.4  & 75.8 & \underline{76.0} & 75.6 & 71.6 & 71.5 & 71.4 & \underline{71.7} & 75.6 & \textbf{76.1} & \textbf{76.1} & \underline{75.4} & 74.9 & 74.9 & 75.5 & 75.6 & \underline{75.9} \\
WResNet & 72.1 & 76.2 & \underline{76.8} & 76.2 & \underline{72.1} & 70.4 & 71.3 & 71.7 & 76.8 & \textbf{77.2} & 76.5 & \underline{75.3} & 74.8 & 75.2 & 76.1 & \underline{76.3} & 76.0 \\
ResNeXt & 75.0 & 78.8 & \underline{79.4} & \underline{79.4}& 74.2 & 71.1 & 73.5 & \underline{74.3} & 79.1 & \textbf{79.9} & 79.3 & \underline{77.6} &77.3 & 76.8 & 77.8 & \underline{79.1} & 78.8 \\
ResNet18 & 70.9 & 77.0 & \textbf{77.4} &  77.1 & \underline{66.3} & 63.7 & 65.3 & 61.9 & 76.1 & \underline{76.4} & 76.0 & 74.8 & \underline{75.6} & 75.9 &76.1 & 76.2 & \underline{76.5} \\ \hline
Mean & 72.9& 76.6 & \underline{76.9} & 76.8 & \underline{71.9} & 70.3 & 71.1 & 70.8 & 76.5 & \textbf{77.1} & 76.6 & 75.6 & \underline{75.7} & 75.6 &76.2 & \underline{76.5} & 76.4 \\ \hline

\end{tabular}%
}
\caption{Clean accuracy values on CIFAR-10 (first 6 rows) and CIFAR-100. \textit{Higher the better}. \underline{Underlined} scores are the best results within their respective group (i.e. single-only, paired-only, etc.). The overall best results are shown in \textbf{bold}. }
\label{tab:cifar10_100_clean_accuracy}
\vspace{-5mm}
\end{table*}

  % clean accuracy
% \usepackage{graphicx}
\begin{table*}[]
\resizebox{\textwidth}{!}{%
\begin{tabular}{l|c||ccc||cccc||cccc||cccc||c}
\multicolumn{1}{c|}{} &  & \multicolumn{3}{c||}{Noise} & \multicolumn{4}{c||}{Blur} & \multicolumn{4}{c||}{Weather} & \multicolumn{4}{c||}{Digital} &  \\
Method & Test Error & Gauss & Shot & Impulse & Defocus & Glass & Motion & Zoom & Snow & Frost & Fog & Brightness & Contrast & Elastic & Pixel & JPEG & mCE \\ \hline
Standard & 23.9 & 79 & 80 & 82 & 82 & 90 & 84 & 80 & 86 & 81 & 75 & 65 & 79 & 91 & 77 & 80 & 80.6 \\
Patch Uniform & 24.5 & 67 & 68 & 70 & 74 & 83 & 81 & 77 & 80 & 74 & 75 & 62 & 77 & 84 & 71 & 71 & 74.3 \\
AA \cite{cubuk2018autoaugment} & 22.8 & 69 & 68 & 72 & 77 & 83 & 80 & 81 & 79 & 75 & 64 & 56 & 70 & 88 & 57 & 71 & 72.7 \\
Random AA \cite{cubuk2018autoaugment} & 23.6 & 70 & 71 & 72 & 80 & 86 & 82 & 81 & 81 & 77 & 72 & 61 & 75 & 88 & 73 & 72 & 76.1 \\
MaxBlur Pool \cite{zhang2019making} & 23.0 & 73 & 74 & 76 & 74 & 86 & 78 & 77 & 77 & 72 & 63 & 56 & 68 & 86 & 71 & 71 & 73.4 \\
SIN \cite{ant_corruption}& 27.2 & 69 & 70 & 70 & 77 & 84 & 76 & 82 & 74 & 75 & 69 & 65 & 69 & 80 & 64 & 77 & 73.3 \\
AugMix \cite{hendrycks2019augmix} & 22.4 & 65 & 66 & 67 & 70 & 80 & 66 & 66 & \textbf{75} & 72 & 67 & 58 & 58 & 79 & 69 & 69 & 68.4 \\
$\mathcal{APR_S}$ \cite{chen2021amplitude} & 24.5 & 61 & 64 & 60 & 73 & 87 & 72 & 81 & 72 & 67 & 62 & 56 & 70 & 83 & 79 & 71 & 70.5 \\
$\mathcal{APR_P}$ \cite{chen2021amplitude}& 24.4 & 64 & 68 & 68 & 70 & 89 & 69 & 81 & 69 & 69 & 55 & 57 & 58 & 85 & 66 & 72 & 69.3 \\
% $\mathcal{APR_{PS}}$ & 24.4 & 55 & 61 & 54 & 68 & 84 & 68 & 80 & 62 & 62 & 49 & 53 & 57 & 83 & 70 & 69 & 65.0 \\ 
$\mathcal{APR_{PS}}$ \cite{chen2021amplitude} & 24.4 & 62 & 68 & 64 & 72 & 86 & 72 & 79 & \textbf{66} & 67 & \textbf{51} & 58 & \textbf{61} & 86 & 66 & 72 & 68.9 \\
\hline
% HA_P & 24.0 & - & - & - & - & - & - & - & - & - & - & - & - & - & - & - & - \\
$\mathcal{HA^{++}_P}$ & \underline{23.5} & 64 & 66 & 67 & 71 & 88 & 72 & 78 & 70 & 69 & 59 & 58 & 64 & 84 & 61 & 69 & 69.7 \\
$\mathcal{HA_{PS}}$ & \textbf{23.2} & 66 & 67 & 62 & 72 & 85 & 77 & 77 & 77 & 71 & 65 & 58 & 69 & 83 & 63 & 69 & 71.2 \\
$\mathcal{HA^{++}_{P}+HA_{S}}$ & 23.8 & 63 & 65 & 60 & 70 & 86 & 71 & 77 & 70 & 68 & \textbf{58} & 58 & 64 & 84 & 62 & 68 & 68.3 \\ 

$\mathcal{HA^{++}_{PS}}$ & 23.7 & \textbf{57} & 61 & 57 & 69 & 85 & 70 & 78 & \textbf{67} & \textbf{66} & \textbf{58} & \textbf{57} & \textbf{63} & 85 & 63 & \textbf{67} & \underline{67.3} \\

$\mathcal{HA^{++}_{PS}}\dagger$ & 25.5 & \textbf{57} & \textbf{58} & \textbf{55} & \textbf{62} & \textbf{75} & \textbf{69} & \textbf{73} & 69 &  68 & 63 & 61 & 68 & \textbf{80} & \textbf{58} & 71 & \textbf{65.8} \\ \hline

PixMix \cite{Hendrycks_2022_CVPR} & \textbf{22.6} & 53 & 52 & 51 & 73 & 88 & 77 & 76 & 62 & 64 & 57 & 56 & 53 & 85 & 69 & 70 & 65.8 \\
DA \cite{hendrycks2021many} & 23.4 & 46 & 47 & 45 & 63 & 75 & 69 & 75 & 67 & 64 & 61 & 55 & 64 & 77 & 50 & 71 & 62.0 \\
DA \cite{hendrycks2021many} + $\mathcal{APR_{PS}}$ & 23.9 & 47 & 48 & 46 & 61 & 73 & 64 & 76 & \textbf{58} & 59 & 53 & 55 & 53 & 77 & 48 & \textbf{68} & 59.1 \\ 
DA \cite{hendrycks2021many} + $\mathcal{HA^{++}_{PS}}$ & 23.9 & 50 & 51 & 47 & 58 & 73 & 62 & 75 & 60 & \textbf{56} & 51 & \textbf{52} & 52 & 77 & 44 & 70 & 58.9 \\
DA \cite{hendrycks2021many} + $\mathcal{HA^{++}_{PS}} \dagger$ & 24.1 & \textbf{45} & \textbf{45} & \textbf{43} & 56 & 69 & 64 & 73 & 61 & 57 & 55 & 53 & 55 & \textbf{74} & \textbf{43} & 76 & 58.1 \\ 
DA \cite{hendrycks2021many} + AM \cite{hendrycks2019augmix}$ + \mathcal{HA^{++}_{PS}}$ & 24.2 & 46 & 47 & 44 & 54 & 73 & \textbf{53} & 67 & 59 & \textbf{56} & \textbf{49} & \textbf{52} & \textbf{50} & 77 & 45 & 73 & \underline{56.4} \\ 
DA \cite{hendrycks2021many} + AM \cite{hendrycks2019augmix}$ + \mathcal{HA^{++}_{PS}}\dagger$ & 24.9 & 46& 46 & 44 & \textbf{52} & \textbf{66} & 54 & \textbf{65} & 59 & 57 & 54 & 53 & 54 & 75 & \textbf{43} & 72 & \textbf{56.1}  \\ \hline

\end{tabular}%
}

\caption{Clean error and corruption robustness on ImageNet. \textit{Lower is better.} The methods shown in the last four rows leverage extra data during training. $\dagger$ indicates training with a higher cut-off frequency.}
\label{tab:imagenet_results}
\end{table*}

  % corruption & clean on imagenet

% \usepackage{graphicx}
% \usepackage[normalem]{ulem}
\begin{table*}[]
\center
\resizebox{0.9\textwidth}{!}{%
\begin{tabular}{l|ccccc|cccccc}
\hline
& AT\cite{wong2020fast} & Cutout\cite{devries2017improved} &  $\mathcal{APR_{P}}$ \cite{chen2021amplitude} & $\mathcal{APR_S}$ \cite{chen2021amplitude} &   $\mathcal{APR_{PS}}$ \cite{chen2021amplitude} & $\mathcal{HA_{S}}$ & $\mathcal{HA^{++}_S}$ & $\mathcal{HA_P}$ & $\mathcal{HA^{++}_P}$ & $\mathcal{HA_{PS}}$ & $\mathcal{HA^{++}_{PS}}$ \\ \hline
CA & 83.3 & 81.3 &  85.3 & 83.5 &  84.4 & \textbf{86.5} & 85.0 & 85.5 & 85.4 & 85.0 & 82.8 \\
RA & 43.2 & 41.6 & 44.0 & 45.0  & 45.4 & 44.1 & 45.4 & 42.1 & 43.5 & 44.8 & \textbf{46.0}  \\ \hline
\end{tabular}%
}
\vspace{1mm}
\caption{Clean and robust accuracy (CA,RA) on CIFAR-10 attacked with AutoAttack \cite{croce2020reliable}. \textit{Higher the better.}}
\label{tab:adversarial_cifar10}
\vspace{-2mm}
\end{table*}

  % AutoAttack

\subsection{Corruption robustness}
\noindent  As mentioned in Section \ref{method_hybridaugment}, we have three augmentation options; $\mathcal{APR}$ \cite{chen2021amplitude}, $\mathcal{HA}$ and $\mathcal{HA^{++}}$. We can apply them using image pairs, a single image or we can do both. This leads to quite a few potential combinations. We now evaluate all these combinations on CIFAR-10 and CIFAR-100, both for clean accuracy and corruption robustness (mCE). 

\noindent \textbf{Comparison against RFC~\cite{mukai2022improving}.} We implement and compare against RFC, which also performs hybrid-image based augmentation. RFC operates on paired-images of same-class samples, therefore we first compare it against $\mathcal{HA_{P}}$ and $\mathcal{HA^{++}_{P}}$. In mCE, we comfortably outperform it while staying competitive in clean accuracy. This shows the value of lifting the limitation of class-based sampling, which RFC does. Note that since we also propose single-image variants, both single-image augmentations ($\mathcal{HA_{S}}$ and $\mathcal{HA^{++}_{S}}$) and combined ones ($\mathcal{HA_{PS}}$ and $\mathcal{HA^{++}_{PS}}$) significantly outperform RFC on all architectures, datasets and metrics. 

\noindent \textbf{Corruption robustness.} The corruption results are shown in Table \ref{tab:cifar10_100_corruption}. The take-away message is crystal clear; $\mathcal{HA^{++}}$ is the best on all datasets, all architectures and all groups. The best results are obtained when we use $\mathcal{HA^{++}}$ both in pairs and single images, further cementing its effectiveness. Note that $\mathcal{HA}$ is competitive or better than $\mathcal{APR}$.

\begin{figure}[!t]
\begin{center}
      \includegraphics[width=0.5\textwidth]{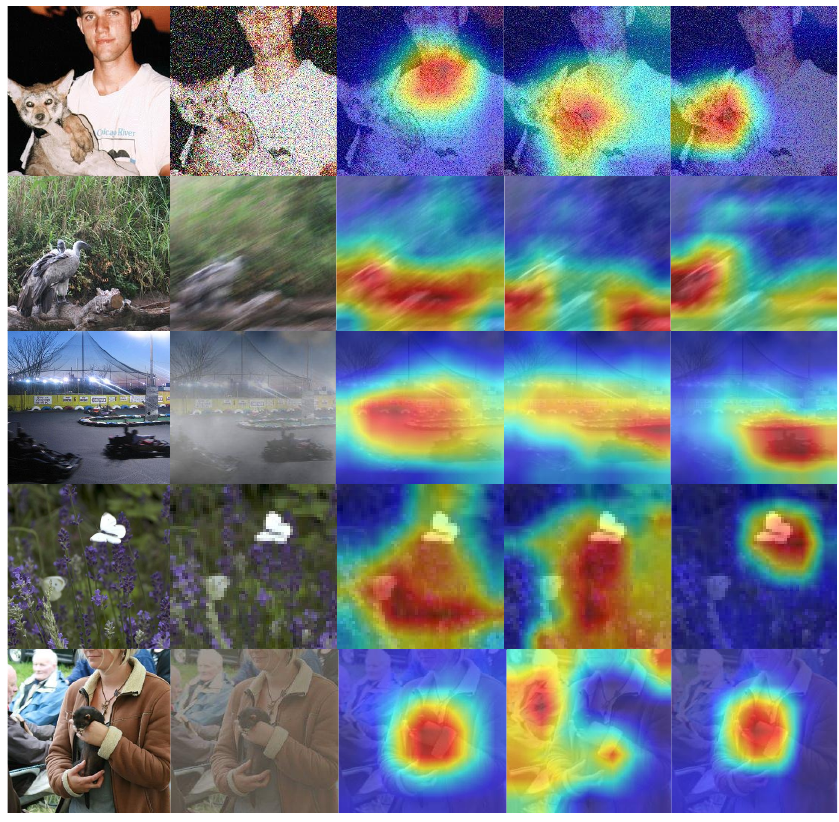}
    \caption{From left to right: ImageNet validation images, their corrupted versions, and Gradcam visualizations \cite{selvaraju2017grad} on standard model, APR \cite{chen2021amplitude} and ours.}    
    \label{fig:gradcam}
\end{center}
\vspace{-8mm}
\end{figure}

\noindent \textbf{Clean Accuracy.} The clean accuracy values of the models shown in Table \ref{tab:cifar10_100_corruption} are given in Table \ref{tab:cifar10_100_clean_accuracy}.  The results show us that both $\mathcal{HA}$ and $\mathcal{HA^{++}}$ achieve a good spot in robustness-accuracy trade-off; except two cases, both of them improve clean accuracy over the original models. The results are not as \textit{clean-cut} as those of Table \ref{tab:cifar10_100_corruption}, but in each group, the best ones mostly include $\mathcal{HA}$ or $\mathcal{HA^{++}}$. Furthermore, the best results on CIFAR-10 and CIFAR-100 have $\mathcal{HA_{S}}$ as the single-image augmentation. Although it does not perform the best, $\mathcal{HA^{++}_{PS}}$ still outperforms the baseline and is highly competitive against others. 

\noindent \textbf{The takeaway.} The results show us that $\mathcal{HA}$ and $\mathcal{HA^{++}}$ are superior to other frequency-based methods, and they comfortably improve robustness and clean accuracy performance across multiple datasets and architectures. See supplementary material for comparison with the state-of-the-art on CIFAR-10 and CIFAR-100. \textit{Hint to readers}: \textit{we achieve the state-of-the-art on all architectures on both datasets}.

\vspace{-3mm}
\subsubsection{Scaling to ImageNet}
\noindent We now assess whether our methods can scale to ImageNet. Since we do not use extra data or ensembles during training or inference, we choose methods with similar characteristics, such as SIN \cite{ant_corruption}, PatchUniform, AutoAugment (AA), Random AA \cite{cubuk2018autoaugment}, MaxBlurPool and AugMix \cite{hendrycks2019augmix}. The results are shown in Table \ref{tab:imagenet_results}. Note that we use pretrained weights for alternative methods if available, otherwise we use the values reported in \cite{chen2021amplitude}.

The results show that all of our variants produce higher clean accuracy compared to $\mathcal{APR}$, showing the value of our method in improving model accuracy. $\mathcal{HA}$ results are competitive in corruption accuracy, but $\mathcal{HA^{++}}$ outperforms both $\mathcal{APR}$ and others in corruption accuracy, while being 0.5 shy of our best clean accuracy. Furthermore, our approach works well with extra data and other augmentations; we apply $\mathcal{HA^{++}_{PS}}$ with DeepAugment \cite{hendrycks2021many} and AugMix \cite{hendrycks2019augmix}, which leads to significant improvements in mCE ($\mathcal{\sim}$11 points) over both DeepAugment and $\mathcal{HA^{++}_{PS}}$.  Note that we are better than $\mathcal{APR}$, even when both methods train with DeepAugment. We also outperform PixMix \cite{Hendrycks_2022_CVPR}, which uses extra training data. Finally, we provide results of $\mathcal{HA^{++}_{PS}}$ with higher cut-off frequency (see experiments with $\dagger$ in Table \ref{tab:imagenet_results}); we see the expected trend where the elimination of higher frequencies make our models more robust in average, at the expense of lowered clean accuracy.

\noindent \textbf{Qualitative results.} We provide GradCam visualizations of $\mathcal{HA^{++}}$ against various corruptions in Figure \ref{fig:gradcam}. We sample corruptions from each category; noise, motion blur, fog, pixelate and contrast corruptions are shown from top to bottom. In the first four rows, it is apparent that corruptions lead to the standard model focusing on the wrong areas, leading to misclassifications. Note that this is the case for $\mathcal{APR}$ as well; it can not withstand these corruptions whereas $\mathcal{HA^{++}}$ still focuses on where matters, and manages to predict correctly. The fifth row shows another failure mode; despite the corruption, standard model manages to predict correctly but $\mathcal{APR}$ loses its focus and leads to misprediction. $\mathcal{HA^{++}}$ does not break what works; this case visualizes the ability of $\mathcal{HA^{++}}$ to improve clean accuracy.

% \usepackage{graphicx}
% \usepackage{graphicx}

\begin{table}[]
\resizebox{\linewidth}{!}{%
\addtolength\tabcolsep{-4pt}
\begin{tabular}{l|cccccc|c}

\multicolumn{1}{c|}{} & \multicolumn{6}{c|}{OOD Datasets} &  \\ \hline
Method & SVHN & LSUN & ImageNet & LSUN$\dagger$ & ImageNet$\dagger$ & CIF100 & Mean \\ \hline
CE & 88.6 & 90.7 & 88.3 & 87.5 & 87.4 & 85.8 & 88.1 \\
CE + CutOut \cite{devries2017improved} & 93.6 & 94.5 & 90.2 & 92.2 & 89.0 & 86.4 & 91.0 \\
CE + Mixup \cite{zhang2017mixup} & 78.1 & 80.7 & 76.5 & 80.7 & 76.0 & 74.9 & 77.8 \\ 
SupCLR \cite{khosla2020supervised} & 97.3 & 92.8 & 91.4 & 91.6 & 90.5 & 88.6 & 92.0 \\
CSI \cite{tack2020csi} & 96.5 & 96.3 & 96.2 & 92.1 & 92.4 & \textbf{90.5} & 94.0 \\ \hline
CE+$\mathcal{APR_S}$ \cite{chen2021amplitude} & 90.4 & 96.1 & 94.2 & 90.9 & 89.1 & 86.8 & 91.3 \\
CE+$\mathcal{APR_P}$ \cite{chen2021amplitude}& \textbf{98.1} & 93.7 & 95.2 & 91.4 & 91.1 & 88.9 & 93.1 \\
CE+$\mathcal{APR_{PS}} \cite{chen2021amplitude}$ & 97.7 & 97.9 & 96.3 & \textbf{93.7} & \textbf{92.8} & 89.5 & \textbf{94.7} \\ \hline
$\mathcal{HA_S}$ & 93.0 & 96.3 & 93.6 & 91.5 & 90.4 & 87.4 & 92.0 \\
$\mathcal{HA_P}$ & 84.9 & 92.8 & 90.0 & 90.5 & 89.1 & 86.9 & 89.0 \\
$\mathcal{HA_{PS}}$ & 95.9 & 97.8 & 95.4 & 91.4 & 90.9 & 87.8 & 93.2 \\
$\mathcal{HA^{++}_P}$ & 92.7 & 92.2 & 91.0 & 89.6 & 89.4 & 86.2 & 90.2 \\
$\mathcal{HA^{++}_S}$ & 94.7 & 97.9 & 96.5 & 91.3 & 89.8 & 86.8 & 92.8 \\
$\mathcal{HA^{++}_{P}}$+$\mathcal{APR_S}$ & 97.5 & \textbf{98.7} & \textbf{97.8} & 93.0 & 91.8 & 89.2 & \textbf{94.7} \\
$\mathcal{HA^{++}_{P}}$+$\mathcal{H_S}$ & 96.9 & 98.3 & 97.1 & 90.6 & 89.9 & 86.4 & 93.2 \\
$\mathcal{HA^{++}_{PS}}$ & 96.6 & 98.7 & 97.7 & 93.0 & 91.2 & 88.1 & 94.2 \\ \hline
\end{tabular}%
}
\caption{Out-of-distribution AUROC results on multiple datasets. \textit{Higher the better}. Our models are trained with CE as well. $\dagger$ indicates fixed versions of respective datasets. CIF100 is CIFAR100.}
\label{tab:ood_results}
\vspace{-5mm}
\end{table}  % out-of-distribution

\subsection{Adversarial Robustness}
We present our results on adversarial robustness in Table \ref{tab:adversarial_cifar10}. For a fair comparison, we train models from scratch if official code is available. If not, we use pretrained models or use the results reported in \cite{chen2021amplitude}. We compare against APR, Cutout and FGSM adversarial training \cite{madry2017towards}.

Our results show that there is no clear winner; with $\mathcal{HA_{S}}$ we obtain the best clean accuracy and with $\mathcal{HA^{++}_{PS}}$ we obtain the best robust accuracy. All our variants are better than the widely accepted adversarial training (AT) baseline in nearly all cases, which shows the effectiveness of our method. Our variants do quite well in clean accuracy and outperform others in nearly all cases. $\mathcal{HA^{++}_{S}}$ offers arguably the best trade-off; it ties with $\mathcal{APR_{PS}}$ on robust accuracy, and outperforms it on clean accuracy.

\subsection{Out-of-Distribution Detection}
For OOD detection, we use a ResNet18 model trained on CIFAR-10 and compare against several configurations, such as training with cross-entropy, SupCLR  \cite{khosla2020supervised} and CSI \cite{tack2020csi}, and augmentation methods as Cutout, Mixup and APR. 

First of all, all our variants comfortably beat the baseline OOD detection (CE), which shows that our proposed method is indeed useful. Furthermore, we see that our proposed methods are highly competitive, and they perform as good as the alternative methods. $\mathcal{HA^{++}_{P} + APR_S}$ outperforms all other methods on LSUN and ImageNet datasets, and produces competitive results on others. Mean AUROC across all datasets show that it ties with the best model $\mathcal{APR_{PS}}$, showing its efficiency. The broader framework we propose leads to many variants with various performance profiles across different datasets, highlighting the flexibility and usefulness of our unification of frequency-centric augmentations. Note that the clean accuracy on CIFAR-10 are provided in Table \ref{tab:cifar10_100_clean_accuracy}, and shows that we perform the same or better than the other methods.

\subsection{Additional results and potential limitations}
\noindent \textbf{ImageNet-$\overline{\text{C}}$.} We also assess our models on ImageNet-$\overline{\text{C}}$ \cite{mintun2021interaction}. The results, given in Table \ref{tab:imagenet_c_bar}, show key insights: we significantly improve over the original standard model and we are just 0.1 shy of $\mathcal{APR_{PS}}$. Training with additional data \cite{hendrycks2021many} helps, and actually puts us ahead of $\mathcal{APR_{PS}}$. An interesting observation is that with higher cut-off frequency (i.e. stronger blur), the performance becomes worse; in ImageNet-C, we observe the opposite. This is potentially due to the different dominant frequency bands in corruptions of ImageNet-C and ImageNet-$\overline{\text{C}}$.

\noindent \textbf{What about transformers?} We also train a Swin-Tiny \cite{liu2021swin} on ImageNet with and without $\mathcal{HA^{++}_{PS}}$; ImageNet-C results show improvements (59.5 vs 54.8 mCE), but at the expense of slight degradation on clean accuracy (81.2 vs 80.6 top-1). Despite the fundamental differences between transformers and CNNs, especially regarding the frequency bands of the features they tend to learn \cite{benz2021robustness}, it is encouraging to see our methods also work well for transformers. We leave further analyses on transformers for future work.

% \usepackage{graphicx}
\begin{table}[]
\resizebox{\linewidth}{!}{%
\addtolength\tabcolsep{-3pt}
\begin{tabular}{l|ccccc|cc}
 & ST &$\mathcal{APR_{PS}}$ & $\mathcal{HA_{PS}}$ & $\mathcal{HA^{++}_{PS}}$ & $\mathcal{HA^{++}_{PS}}\dagger$  &$\mathcal{APR_{PS}}\ddagger$ &$\mathcal{HA^{++}_{PS}}\ddagger$  \\ \hline
Error & 61.0 &52.1 & 56.2 & 52.2 & 53.4 & 48.6 & \textbf{47.9} 
\end{tabular}%
}
\caption{Error values on ImageNet-$\overline{\text{C}}$. $\dagger$ indicates training with a higher cut-off frequency. $\ddagger$ indicates training with DeepAugment\cite{hendrycks2021many}. \textit{ST} indicates standard model training.}
\label{tab:imagenet_c_bar}
\vspace{-5mm}
\end{table}

\section{Conclusion} \label{sec:hybrid_augment_conclusion}
In this paper, inspired by the frequency-centric explanations of how CNNs generalize, we propose two augmentations methods \textit{HybridAugment} and \textit{HybridAugment++}. The former aims to reduce the reliance of CNN generalization on high-frequency information in images, whereas the latter does the same but also promotes the use of phase information rather than the amplitude component. This unification of two distinct frequency-based analyses into a data augmentation method leads to results competitive to or better than state-of-the-art on clean accuracy, corruption and adversarial performance and out-of-distribution detection.

{\small
\bibliographystyle{ieee_fullname}
\bibliography{egbib}
}

\clearpage
\section{Supplementary Material}

\begin{table}[b!]
 \resizebox{\textwidth}{!}{%
 \addtolength\tabcolsep{-3pt}
 \begin{tabular}{l|ccccccc||ccc||ccc||ccc||cc}
 & \multicolumn{7}{c||}{State-of-the-art methods} & \multicolumn{3}{c||}{Single-only} & \multicolumn{3}{c||}{Paired-only} & \multicolumn{3}{c||}{Combined} & \multicolumn{2}{c}{$\mathcal{APR_{P}}$\cite{chen2021amplitude} $with$}  \\
   & Orig & Cutout  & Mixup & CutMix & AT & AugMix & AA & $\mathcal{APR_{S}}$ & $\mathcal{HA_{S}}$ & $\mathcal{HA^{++}_{S}}$ & $\mathcal{APR_{P}}$ & $\mathcal{HA_{P}}$ & $\mathcal{HA^{++}_{P}}$ & $\mathcal{APR_{PS}}$ & $\mathcal{HA_{PS}}$ & $\mathcal{HA^{++}_{PS}}$ & $\mathcal{HA_{S}}$ & $\mathcal{HA^{++}_{S}}$ \\ \hline
  AllConv & 30.8 & 32.9 & 24.6 & 31.3 & 28.1 & 15.0 & 29.2 & 14.8 & 16.8 & 13.9 & 21.5 & 20.8 & 16.7 & 11.5 & 12.0 & \textbf{10.7} & 11.9 & 11.2 \\
    DenseNet & 30.7 & 32.1 & 24.6 & 33.5 & 27.6 & 12.7 & 26.6 & 12.3 & 15.0 & 11.1 & 20.3 & 18.4 & 14.2 & 10.3 & 10.9 & \textbf{9.5} & 10.6 &  10.2 \\
  WResNet & 26.9 & 26.8 & 22.3 & 27.1 & 26.2 & 11.2 & 23.9 & 10.6 & 13.6 & 10.0 & 18.3 & 16.4 & 13.2 & 9.1 & 9.9 & \textbf{8.3} &9.2 & 8.7 \\
   ResNeXt & 27.5 & 28.9 & 22.6 & 29.5 & 27.0 & 10.9 & 24.2 & 11.0 & 13.2 & 9.9 & 18.5 & 17.6 & 13.2 & 9.1 & 10.3 & \textbf{7.9} & 9.3 & 8.7 \\ \hline
   % ResNet18 & - & - & - & - & - & - & - & 9.9 & 12.2 & 9.34 & 17.0 & 18.3 & 15.2 & 9.1 & 9.3 & \textbf{8.2} & 9.0 & 8.5 \\ \hline
    Mean & 29.0 & 30.2 & 23.5 & 30.3 & 27.2 & 12.5 & 26.0 & 12.1 & 14.6 & 11.2 & 19.6 & 18.3 & 14.3 & 10.0 & 10.7 & \textbf{9.1} & 10.2 & 9.7 \\ \hline
  AllConv & 56.4 & 56.8 & 53.4 & 56.0 & 56.0 & 42.7 & 55.1 & 39.8 & 43.0 & 38.9 & 47.5 & 44.7 & 41.7 & 35.9 & 36.5 & \textbf{34.4} & 35.9 & 35.1 \\
   DenseNet & 59.3 & 59.6 & 55.4 & 59.2 & 55.2 & 39.6 & 53.9 & 38.3 & 41.3 & 37.3 & 49.8 & 45.6 & 41.8 & 35.8 & 36.1 & \textbf{33.4} & 36.3 & 35.0 \\
   WResNet & 53.3 & 53.5 & 50.4 & 52.9 & 55.1 & 35.9 & 49.6 & 35.5 & 38.1 & 33.9 & 44.7 & 43.1 & 39.3 & 32.9 & 34.2 & \textbf{31.2} & 33.2 & 31.9 \\
   ResNeXt & 53.4 & 54.6 & 51.4 & 54.1 & 54.4 & 34.9 & 51.3 & 33.7 & 35.6 & 31.1 & 44.2 & 41.2 & 36.4 & 31.0 & 31.5 & \textbf{28.8} & 31.2 & 29.9 \\ \hline
   % ResNet18 & - & - & - & - & - & - & - & 33.0 & 35.6 & 32.1 & 49.2 & 45.5 & 44.6 & 31.8 & 31.8 & \textbf{29.9} & 32.5 & 31.2 \\ \hlin    
    Mean & 55.6 & 56.1 & 52.6 & 55.5 & 55.2 & 38.3 & 52.5 & 36.8 & 39.5 & 35.3 & 46.5 & 43.6 & 39.8 & 33.9 & 34.5 & \textbf{31.9}  & 34.1 &  33.0\\ \hline
\end{tabular}%
 }
 \parbox{\textwidth}{\caption{Corruption robustness on CIFAR-10 (first 6 rows) and CIFAR-100 with various CNNs. Values show mCE, \textit{lower is better}. The table is divided into groups for easy comparison; single-only augmentation, paired-only augmentation, combined augmentations, etc. \textit{Orig} refers to the standard model.}}
 \label{tab:supmat_table1}
 \vspace{-3mm}
 \end{table}

 \begin{table}[b!]
 \resizebox{\textwidth}{!}{%
 \addtolength\tabcolsep{-3pt}
 \begin{tabular}{l|ccccccc||ccc||ccc||ccc||cc}
 & \multicolumn{7}{c||}{State-of-the-art methods} & \multicolumn{3}{c||}{Single-only} & \multicolumn{3}{c||}{Paired-only} & \multicolumn{3}{c||}{Combined} & \multicolumn{2}{c}{$\mathcal{APR_{P}}$\cite{chen2021amplitude} $with$}  \\ 
    & Orig & Cutout  & Mixup & CutMix & AT & AugMix & AA & $\mathcal{APR_{S}}$ & $\mathcal{HA_{S}}$ & $\mathcal{HA^{++}_{S}}$ & $\mathcal{APR_{P}}$ & $\mathcal{HA_{P}}$ & $\mathcal{HA^{++}_{P}}$ & $\mathcal{APR_{PS}}$ & $\mathcal{HA_{PS}}$ & $\mathcal{HA^{++}_{PS}}$ & $\mathcal{HA_{S}}$ & $\mathcal{HA^{++}_{S}}$ \\ \hline
 AllConv & 93.9 & 93.9 & 93.7 & 93.6 & 81.1 & 93.5 & 93.5 & 93.5 & 94.1 & 93.9 & \textbf{94.5} & 93.9 & 94.0 & 94.3 & \textbf{94.5} & 94.3 & 94.3 & 94.3 \\
 DenseNet & 94.2 & \textbf{95.2} & 94.5 & 94.7 & 82.1 & 95.1 & 95.2 & 94.9 & 94.7 & 95.0 & 95.0 & 93.1 & 93.2 & \textbf{95.2} & 94.9 & 94.8 & 95.1 & 95.1 \\
 WResNet & 94.8 & 95.6 & 95.1 & 95.4 & 82.9 & 95.1 & 95.2 & 95.0 & 95.3 & 95.4 & 95.2 & 93.2 & 92.0 & 95.7 & 95.0 & 95.3 & 95.4 & \textbf{95.8} \\
 ResNeXt & 95.7 & 95.6 & 95.8 & 96.1 & 84.6 & 95.8 & 96.2 & 95.5 & 95.3 & 95.7 & 95.5 & 93.5 & 92.9 & \textbf{96.1} & 95.2 & 95.9 & 95.6 & \textbf{96.1} \\ \hline
 Mean & 94.2 & 95.0 & 94.7 & 94.9 & 82.6 & 94.8 & 95.0 & 94.9 & 94.9 & 95.1 & 95.0 & 92.9 & 92.3 & 95.2 & 95.0 & 95.1 & 95.1 & \textbf{95.3} \\ \hline
 AllConv & 74.9 & - & - & - & - & - & - & 75.3 & 75.0 & \textbf{75.8} & 74.8 & 74.08 & 74.7 & 75.2 & \textbf{75.8} & 75.2 & 75.7 & 75.1 \\
 DenseNet & 71.4 & - & - & - & - & - & - & 75.8 & 76.0 & 75.6 & 71.5 & 71.4 & 71.7 & 75.6 & 74.9 & 75.9 & \textbf{76.1} & \textbf{76.1} \\
 WResNet & 72.1 & - & - & - & - & - & - & 76.2 & 76.8 & 76.2 & 70.4 & 71.3 & 71.7 & 76.8 & 74.8 & 76.0 & \textbf{77.2} & 76.5 \\
 ResNeXt & 75.0 & - & - & - & - & - & - & 78.8 & 79.4 & 79.4 & 71.1 & 73.5 & 74.3 & 79.1 & 77.3 & 78.8 & \textbf{79.9} & 79.3 \\ \hline
 Mean & 72.9 & - & - & - & - & - & - & 76.6 & 76.9 &  76.8 & 70.3 & 71.1 & 70.8 & 76.5 & 75.6 & 76.4 & \textbf{77.1} & 76.6 \\ \hline
 \end{tabular}%
 }
  \parbox{\textwidth}{\caption{Clean accuracy values on CIFAR-10 (first 6 rows) and CIFAR-100. \textit{Higher the better}. The table is divided into groups for easy comparison; single-only augmentation, paired-only augmentation, combined augmentations, etc.\textit{Orig} refers to the standard model.}}
\label{tab:suppmat_tab_2}
\vspace{-5mm}
\end{table}

\subsection{State of the art comparison on CIFAR-C}

In the main text, we provide detailed ablations on CIFAR10/100-C in the form of corruption robustness evaluation. Due to space limitations, we could not provide state-of-the-art results there; we provide these results here. We compare ourselves with methods which share our characteristics; no additional data or models to be used. We choose CutOut \cite{devries2017improved}, Mixup \cite{zhang2017mixup}, CutMix \cite{yun2019cutmix}, adversarial training (AT) \cite{madry2017towards}, AutoAugment (AA) \cite{cubuk2018autoaugment}, Augmix \cite{hendrycks2019augmix} and APR \cite{chen2021amplitude}. We take the results of these methods from \cite{chen2021amplitude}; we do not include CIFAR-100 clean accuracy results or ResNet18 results here since they are not available.

\noindent \textbf{Corruption Robustness.} Table \ref{tab:supmat_table1} shows mCE values of other methods, as well as the best results provided in Table 1 of the main text. The inclusion of the state-of-the-art methods do not change the takeaway message; $\mathcal{HA^{++}_{PS}}$ comfortably outperforms others on all datasets and architectures. Note that all variants of $\mathcal{HA}$ and $\mathcal{HA^{++}}$ either outperform or are competitive to all state-of-the-art methods.

\noindent \textbf{Clean Accuracy.} Table \ref{tab:suppmat_tab_2} shows clean accuracy values of other methods, as well as the best results provided in Table 2 of the main text. $\mathcal{HA^{++}_{PS}}$ outperforms all other state-of-the-art methods, and the best CIFAR-10 result comes with $\mathcal{APR_{P}} + \mathcal{HA^{++}_{S}}$.  Note that the best result on CIFAR-100 comes with $\mathcal{APR_{P}} + \mathcal{HA_{S}}$, which shows the effectiveness of our proposed methods.

\subsection{More on $\mathcal{HA}$ and $\mathcal{HA^{++}}$}
We provide the pseudo-code of $\mathcal{HA^{++}_{P}}$ and $\mathcal{HA_{P}}$ in Algorithm \ref{fig:pseudo_code}. Also provided is the pseudo-code for $\mathcal{HA^{++}_{S}}$ and $\mathcal{HA_{S}}$ in Algorithm \ref{fig:pseudo_code_2}. Our code and pretrained models will be made publicly available.

Note that in Algorithm \ref{fig:pseudo_code_2}, we decompose into low and high frequency bands both augmented images (lines 22-23 and 25-26), and also amplitude-phase swap low-frequency bands (\verb|lfc_f| and \verb|lfc_s|) of both augmented images (lines 42 and 56). We then randomize the selection of which low/high frequency components will come from which image for the final result (lines 58 to 63). Figure 1 of the main text shows a simplified version of this, where only the execution of line 61 is shown. In practice, we use the implementation provided in Algorithm \ref{fig:pseudo_code_2}.

\subsection{Detailed results - transformer}
We provide the detailed results of our corruption robustness experiments with Swin-Tiny \cite{liu2021swin}. The result in Table \ref{tab:suppmat_swin}
 shows that $HA^{++}_{PS}$ consistently improves on all types of corruptions, regardless of their frequency characteristics. 

\clearpage

\renewcommand{\lstlistingname}{Code} % Code to Algorithm
\begin{lstlisting}[language=python,caption={PyTorch-style pseudocode for $\mathcal{HA_{S}}$ and $\mathcal{HA^{++}_{S}}$.},captionpos=b,frame=single, label={fig:pseudo_code_2}]
def hybrid_augment_single(x, prob, blur_fnc, sample_augs, is_ha_plus):
    #x: a single training image
    #prob: probability value [0,1]
    #blur_fnc: blurring function
    #sample_augs: randomly sample augmentations
    #is_ha_plus: True for HA++, false for HA
    #fft, ifft: fourier and inverse fourier transform 
    
    p = random.uniform(0,1)
    if p > prob:
        return x

    #First augmented view.
    ops1 = sample_augs()
    x_aug1 = ops1(x)

    #Second augmented view.
    ops2 = sample_augs()
    x_aug2 = ops2(x)

    lfc_f = blur_fnc(x_aug1)
    hfc_f  = x_aug1 - lfc_f

    lfc_s = blur_fnc(x_aug2)
    hfc_s  = x_aug2 - lfc_s

    if is_ha_plus:
        #For lfc_f.
        p = random.uniform(0, 1)
        if p > 0.6:
            lfc_f = lfc_f
        else:
            ops3 = sample_augs()
            lfc_aug = ops3(lfc_f)
            ops4 = sample_augs()
            lfc_aug_2 = ops4(lfc_f)
            
            phase1, amp1 = fft(lfc_aug)
            phase2, amp2 = fft(lfc_aug_2)
            lfc_f = ifft(phase1, amp2)

        #For lfc_s.
        p = random.uniform(0, 1)
        if p > 0.6:
            lfc_s = lfc_s
        else:
            ops5 = sample_augs()
            lfc_aug = ops5(lfc_s)
            ops6 = sample_augs()
            lfc_aug_2 = ops6(lfc_s)
            
            phase1, amp1 = fft(lfc_aug)
            phase2, amp2 = fft(lfc_aug_2)
            lfc_s = ifft(phase1, amp2)
    
    p = random.uniform(0, 1)
    
    if p > self.prob:
        hybrid_im = lfc_f + hfc_s
    else:
        hybrid_im = lfc_s + hfc_f

    return hybrid_im
\end{lstlisting}
\renewcommand{\lstlistingname}{Code} % Code to Algorithm

% \begin{figure}
% \begin{minted}[breaklines, linenos, breakafter=d,fontsize=\footnotesize,frame=lines]{python}
% def hybrid_augment_paired(x_batch, prob, blur_fnc, is_ha_plus):
%     #x_batch: batch of training images
%     #prob: probability value [0,1]
%     #blur_fnc: blurring function
%     #is_ha_plus: True for HA++, false for HA
%     #fft: fourier transform 
%     #ifft: inverse fourier transform 
    
%     p = random.uniform(0,1)
%     if p > prob:
%         return x

%     batch_size = x_batch.size()[0]
%     index = torch.randperm(batch_size)

%     lfc = blur_fnc(x_batch)
%     hfc = x - lfc
%     hfc_mix = hfc[index]

%     if is_ha_plus:
%         #Based on the APR method.
%         p = random.uniform(0,1)
%         if p > 0.6:
%             lfc  = lfc
%         else:
%             index_p = torch.randperm(batch_size)
%             phase1, amp1 = fft(lfc)
%             lfc_mix = lfc[index_p]
%             phase2, amp2 = fft(lfc_mix)
%             lfc = ifft(phase1, amp2)
    
%     hybrid_ims = lfc + hfc_mix
%     return hybrid_ims
% \end{minted}
% \vspace{-3mm}
% \caption{PyTorch-style pseudocode for $\mathcal{HA_{P}}$ and $\mathcal{HA^{++}_{P}}$.}
% \label{fig:pseudo_code}
% \vspace{-3mm}
% \end{figure}

\begin{lstlisting}[language=python,caption={PyTorch-style pseudocode for $\mathcal{HA_{P}}$ and $\mathcal{HA^{++}_{P}}$.},captionpos=b, frame=single,label={fig:pseudo_code}]
def hybrid_augment_paired(x_batch, prob, blur_fnc, is_ha_plus):
    #x_batch: batch of training images
    #prob: probability value [0,1]
    #blur_fnc: blurring function
    #is_ha_plus: True for HA++, false for HA
    #fft: fourier transform 
    #ifft: inverse fourier transform 
    
    p = random.uniform(0,1)
    if p > prob:
        return x

    batch_size = x_batch.size()[0]
    index = torch.randperm(batch_size)

    lfc = blur_fnc(x_batch)
    hfc = x - lfc
    hfc_mix = hfc[index]

    if is_ha_plus:
        #Based on the APR method.
        p = random.uniform(0,1)
        if p > 0.6:
            lfc  = lfc
        else:
            index_p = torch.randperm(batch_size)
            phase1, amp1 = fft(lfc)
            lfc_mix = lfc[index_p]
            phase2, amp2 = fft(lfc_mix)
            lfc = ifft(phase1, amp2)
    
    hybrid_ims = lfc + hfc_mix
    return hybrid_ims
\end{lstlisting}
\renewcommand{\lstlistingname}{Code} % Code to Algorithm

\begin{table*}[ht!]
\resizebox{\textwidth}{!}{%
 \addtolength\tabcolsep{-3pt}
\begin{tabular}{l|c||ccc||cccc||cccc||cccc||c}
\multicolumn{1}{c|}{} &  & \multicolumn{3}{c||}{Noise} & \multicolumn{4}{c||}{Blur} & \multicolumn{4}{c||}{Weather} & \multicolumn{4}{c||}{Digital} &  \\
Method & Test Error & Gauss & Shot & Impulse & Defocus & Glass & Motion & Zoom & Snow & Frost & Fog & Brightness & Contrast & Elastic & Pixel & JPEG & mCE \\ \hline
Standard & \textbf{18.8}  & 52 & 54 & 53 &  68 & 81 & 65 & 72 & 57 & 52 & 47 & 48 & 45 & 74 & 61 & 63  & 59.5 \\

$\mathcal{HA^{++}_{PS}}$ & 19.4 & \textbf{44} & \textbf{48} & \textbf{42} & \textbf{63} & \textbf{78} & \textbf{59} & \textbf{71} & \textbf{49} & \textbf{48} & \textbf{46} & \textbf{46} & \textbf{39} & \textbf{71} & \textbf{60} & \textbf{59} & \textbf{54.8}\\ \hline

\end{tabular}%
}
\parbox{\textwidth}{\caption{Swin-Tiny Clean error and corruption robustness (mCE) on ImageNet. \textit{Lower is better.}}}
\label{tab:suppmat_swin}
\end{table*}

\subsection{Related work continued}
The robustness literature is vast, and it is difficult to cover all methods, therefore in the main text we opted to cover and compare ours against the most relevant ones (i.e. frequency-centric augmentations). Here, we discuss additional, more recent methods.

We focus on recent methods, such as \cite{calian2022defending,wang2021augmax,Saikia_2021_ICCV,prime_aug}. \cite{calian2022defending} uses an extra model to generate new training samples, which makes the method significantly more complex than ours. Despite this added complexity, we outperform it on ImageNet-C without extra data (75.03 vs 65.8 mCE) and with extra data (62.9 vs 58.9 mCE), even though they use additional augmentations (i.e. AugMix). \cite{wang2021augmax} extends AugMix by making parts of the cascade augmentation pipeline learnable. We outperform it on CIFAR-10/100-C on all architectures. Note that we could not compare against them  on ImageNet-C as they use a different architecture (i.e. ResNet18). \cite{Saikia_2021_ICCV} outperforms us on ImageNet, but it uses model ensembles during training, which are finetuned on some of the test-time corruptions of ImageNet-C (i.e. noise and blur finetuning for high-frequency model, contrast finetuning for low-frequency model). We believe this violates the assumption of not using test-time corruptions in training. PRIME \cite{prime_aug} mixes several max-entropy transforms to augment the training distribution. We outperform it on CIFAR-10/100, are competitive on ImageNet-$\bar{C}$ but behind on ImageNet-C. Despite its results, PRIME has three key disadvantages compared to our method; it i) requires per-dataset hyperparameter tuning for its transforms, ii) manual tuning of these parameters are required to preserve semantics after augmentation and iii) shows that their augmented images look similar to test-time corruptions, which might be (inadvertently) violating the assumption of not using test-time corruptions in training.

\subsection{Adversarial robustness on ImageNet}
We evaluate ResNet-50 models trained with $\mathcal{HA^{++}_{PS}}$, $\mathcal{APR_{PS}}$ and standard training.  We use the model checkpoints shown in Table \ref{tab:imagenet_results}; we do not train new models. Table \ref{tab:adv_imagenet} shows $\mathcal{HA^{++}_{PS}}$ improves robust and clean accuracy (RA, CA) on ImageNet, and comfortably outperforms our baseline. Note that we use a smaller $\epsilon=1/255$ value, as higher epsilon evaluation would require adversarial (re)training. 

\begin{table}[h]
\centering
\resizebox{0.2\textwidth}{!}{%
\setlength\tabcolsep{1.5pt} 
\begin{tabular}{c|c|c|c}
 & Orig. &  $\mathcal{APR_{PS}}$ & $\mathcal{HA^{++}_{PS}}$   \\  \hline
CA & 76.10 & 75.60  & \textbf{76.30} \\
RA & 51.02 &  54.22  & \textbf{56.44} 
\end{tabular}%
}
\caption{AutoAttack results.}
\label{tab:adv_imagenet}
\end{table}

\subsection{Transfer learning performance}

As reported in \cite{salman2020adversarially}, robust models tend to transfer better to downstream tasks. In the same vein, we perform a wide range of finetuning experiments, where a standard ResNet50 and $\mathcal{HA^{++}_{PS}}$-trained ResNet50 are finetuned on various datasets by changing the final layer. Note that we do not train new models; we use the model checkpoints shown in Table \ref{tab:imagenet_results}. Table \ref{tab:transfer_learning} shows we comfortably outperform standard training on majority of other classification tasks. This shows the transferability of the features learned by our augmentation schemes.
  
\vspace{7mm}
\begin{table}[h]
\centering
\resizebox{0.45\textwidth}{!}{%

\begin{tabular}{c|c|c|c|c|c|c|c|c|c|c|c}
 \begin{rotate}{60} CIFAR10 \end{rotate} 
 & \begin{rotate}{60} CIFAR100 \end{rotate}  
 & \begin{rotate}{60} Aircraft \end{rotate} 
 & \begin{rotate}{60} CTech101 \end{rotate} 
 & \begin{rotate}{60} DTD \end{rotate} 
 & \begin{rotate}{60} Flowers \end{rotate} 
 & \begin{rotate}{60} Pets \end{rotate} 
 & \begin{rotate}{60} CTech256 \end{rotate}  
 & \begin{rotate}{60} Birds \end{rotate} 
 & \begin{rotate}{60} Cars \end{rotate}  
 & \begin{rotate}{60} SUN \end{rotate}  
 & \begin{rotate}{60} Food \end{rotate} \\ \hline

 96.8 & 83.4 & \textbf{86.6} & \textbf{94.0} & 74.1 & 96.3 & 93.2 & 81.5 & 73.6 & 90.9 & 62.1 & \textbf{87.5} \\ 
 \textbf{97.4} & \textbf{84.9} & 84.5 & 92.7 & \textbf{75.1} & \textbf{96.8} & \textbf{93.3} & \textbf{83.0} & \textbf{73.7} & \textbf{91.0} & \textbf{63.3} & 87.4
\end{tabular}%
}
\caption{Transfer learning acc. (top-1) of standard  ResNet50 (top) and $\mathcal{HA^{++}_{PS}}$ (bottom) on 12 other classification datasets. }
\label{tab:transfer_learning}
\vspace{-1mm}
\end{table}
% \end{wraptable}

\end{document}